\documentclass[a4paper,fleqn]{cas-sc}

\usepackage[numbers]{natbib}

\def\tsc#1{\csdef{#1}{\textsc{\lowercase{#1}}\xspace}}
\tsc{WGM}
\tsc{QE}
\tsc{EP}
\tsc{PMS}
\tsc{BEC}
\tsc{DE}

\begin{document}
\let\WriteBookmarks\relax
\def\floatpagepagefraction{1}
\def\textpagefraction{.001}
\shorttitle{Autonomous Lane Detection}
\shortauthors{M. Tangestanizadeh et~al.}

\title [mode = title]{Attention-based U-Net Method for Autonomous Lane Detection}                      



\author[1]{Mohammadhamed Tangestanizadeh}[orcid=0009-0004-9703-6559]
\cormark[1]
 \ead{mtangest@ucsc.edu}


\affiliation[1]{organization={University Of California},
                addressline={1156 High St}, 
                city={Santa Cruz},
                postcode={95064}, 
                state={California},
                country={USA}}

\author[2]{Mohammad {Dehghani Tezerjani}}[orcid=0009-0003-0790-9831]
\ead{mike.degany@unt.edu}

\author[2]{Saba {Yousefian Jazi}}[orcid=0009-0001-8665-0797]
\ead{sabayousefianjazi@my.unt.edu}


\affiliation[2]{organization={University Of North Texas},
                addressline={2204 W Prairie St}, 
                postcode={76201},  
                city={Denton},
                state={Texas},
                country={USA}}


\cortext[cor1]{Corresponding author}


\begin{abstract}
Lane detection involves identifying lanes on the road and accurately determining their location and shape. This is a crucial technique for modern assisted and autonomous driving systems. However, several unique properties of lanes pose challenges for detection methods. The lack of distinctive features can cause lane detection algorithms to be confused by other objects with similar appearances. Additionally, the varying number of lanes and the diversity in lane line patterns, such as solid, broken, single, double, merging, and splitting lines, further complicate the task. To address these challenges, Deep Learning (DL\footnotemark[1]) \fntext[fn1]{Deep Learning} approaches can be employed in various ways. Merging DL models with an attention mechanism has recently surfaced as a new approach. In this context, two deep learning-based lane recognition methods are proposed in this study. The first method employs the Feature Pyramid Network (FPN\footnotemark[2]) \fntext[fn2]{Feature Pyramid Network} model, delivering an impressive 87.59\% accuracy in detecting road lanes. The second method, which incorporates attention layers into the U-Net model, significantly boosts the performance of semantic segmentation tasks. The advanced model, achieving an extraordinary 98.98\% accuracy and far surpassing the basic U-Net model, clearly showcases its superiority over existing methods in a comparative analysis. The groundbreaking findings of this research pave the way for the development of more effective and reliable road lane detection methods, significantly advancing the capabilities of modern assisted and autonomous driving systems.    
\end{abstract}




\maketitle

\section{Introduction}
Lanes are critical traffic signs that divide roads, ensuring automobiles drive safely and efficiently \cite{RN13}. Lane detection involves automatically identifying road markers to keep cars within their assigned lanes and prevent collisions with vehicles in other lanes, playing a key role in autonomous driving \cite{RN14}. Accurate lane detection enables autonomous vehicles to make informed decisions about their position and status, ensuring safe driving \cite{RN15}. However, lane detection is challenging due to the variety of lane markers, complex and changing road conditions, and the inherent slenderness of lanes \cite{RN16}. Consequently, extensive research has been dedicated to developing reliable lane-detection algorithms \cite{RN11}. Lane detection also plays a crucial role in trajectory planning, helping vehicles navigate complex environments, especially when obstacles are present, requiring precise path adjustments to ensure safe and efficient vehicle operation \cite{tezerjani2024real}. 
Traditionally, lane detection relied on hand-crafted methods, including geometric modeling and conventional approaches, which involved image pre-processing, feature extraction, lane model fitting, and line tracking. These traditional methods were often complex and time-consuming due to their manual nature \cite{RN17}. In recent years, the advent of Artificial Intelligence (AI) technologies has significantly improved various autonomous vehicle components \cite{tezerjani2024survey}, including lane marking recognition, making it more accessible, faster, and efficient without the need for hand-crafted procedures. AI, particularly Machine Learning (ML) and Deep Learning (DL), has revolutionized lane detection \cite{RN18}. DL has gained popularity over ML for its superior performance in classification and detection tasks using image frames as input to the network algorithm \cite{RN19}. While some researchers advocate using DL as a standalone approach, others recommend integrating it with additional methods to enhance its effectiveness in challenging conditions. Recently, a new integration approach has emerged, combining DL with an attention mechanism to improve lane detection accuracy and reliability further \cite{RN20}.
\\This paper introduces two innovative deep-learning methods for line recognition. Our contributions include the application of the Feature Pyramid Network (FPN) method to enhance image feature extraction and line detection. Additionally, we integrate an attention layer with the U-Net model, significantly boosting the performance of semantic segmentation tasks. The exceptional performance of our attention-based U-Net model, which outperforms similar models in the IoU criterion, underscores its potential to revolutionize line detection technology.
\\The structure of this paper is as follows: The II Section reviewed existing studies in the field of lane detection. The Detailed implementation procedures for the two proposed methods are provided in the III Section. The IV Section evaluated the performance of the FPN model and the attention-based U-Net model using established criteria, demonstrating the superior performance of both models. Finally, the findings of this research are summarized, and suggestions for future research directions are offered in the V Section.
\section{Related works}
In recent years, lane detection has emerged as a critical area of research within the scientific community. Researchers around the world are actively developing innovative methods for accurately detecting road lanes, resulting in a wealth of methodologies documented in technical literature. Notably, deep learning-based approaches have gained prominence, demonstrating significant effectiveness in addressing the challenges associated with lane detection.
\newline Zhang et al. \cite{RN1} introduced a hierarchical Deep Hough Transform (DHT) strategy referred to as HoughLaneNet, which integrates all lane characteristics within an image into the Hough parameter space. The approach proposed by the researchers enhances the point selection technique and introduces a Dynamic Convolution Module to distinguish between lanes effectively in the original image. The architecture of their neural network includes a backbone network, such as a ResNet or Pyramid Vision Transformer, a Feature Pyramid Network serving as the intermediary component for extracting multi-scale features, and a hierarchical DHT-based feature aggregation head for precise lane segmentation.
\\Khan et al. \cite{RN2} have developed an innovative CNN model called LLDNet, which features a lightweight encoder-decoder architecture. LLDNet excels at estimating lanes in deteriorated road conditions, as well as under adverse circumstances.
\\Andrei et al.\cite{RN3} enhanced a pre-existing lane detection method by modifying two key features, resulting in improved optimization and reduced false lane marker detection. They proposed replacing the standard Hough transform with a probabilistic Hough transform and utilizing a parallelogram Region Of Interest (ROI) instead of a trapezoidal one. These changes led to a runtime increase of approximately 30\% and an accuracy improvement of up to 3\% compared to the original method.
\\Muthalagu et al. \cite{RN4} introduced two advanced methods for lane and object detection, designed to handle diverse and complex traffic scenarios, especially where driving speeds are high. The images captured from the model car are processed, and the processed data is used to control the autonomous vehicle. The first method is a minimalistic lane detection approach that identifies only straight lane lines. The second method leverages a CNN-based model that learns to drive by analyzing driver behavior data, allowing the autonomous vehicle to mimic human driving patterns.
\\Sapkal et al. \cite{RN5} developed a lane detection system utilizing both U-Net and Segnet models, which they tested on the Tusimple dataset. The U-Net model demonstrated superior performance, achieving an accuracy of 97.52\%, surpassing that of the Segnet model.
\\Honda et al. \cite{RN6} introduced LaneIoU which better correlates with lane detection accuracy by considering local lane angles. Also, they developed a novel detector named CLRerNet, which incorporates LaneIoU into the target assignment cost and loss functions, aiming to enhance the quality of confidence scores.
\\Saika et al.\cite{RN7} created a real-time vehicle and lane detection system based on a modified version of the OverFeat CNN. They created a comprehensive highway dataset comprising 17,000 picture frames with vehicle bounding boxes and over 616,000 image frames with lane annotations, collected using cameras, lidar, radar, and GPS. This data was used to train a CNN architecture capable of recognizing all lanes and vehicles in a single forward pass.
\\Dewangan et al. \cite{RN8} proposed a semantic segmentation architecture utilizing an encoder-decoder network. Their hybrid model, combining U-Net and ResNet, begins by down-sampling the image and extracting essential features with ResNet-50. Then, U-Net up-samples and decodes the image segments based on these detected features. 
\\Lee and Liu \cite{RN9}, inspired by the U-Net architecture for semantic image segmentation, proposed a lightweight version called DSUNet, which uses depthwise separable convolutions for end-to-end lane detection and Path Prediction (PP) in autonomous driving. They also developed and integrated a PP algorithm with a CNN to create a simulation model (CNN-PP). This model can qualitatively, quantitatively, and dynamically assess CNN’s performance in real-time autonomous driving, with the host agent car driving alongside other agents. DSUNet is 5.12 times lighter in model size and 1.61 times faster in inference than U-Net.
\\Munir et al. \cite{RN11} combined a deep learning algorithm with an attention mechanism to enhance road lane detection. Their proposed Lane Detection Network (LDNet) employs an encoder-decoder structure with an Atrous Spatial Pyramid Pooling (ASPP) block and an attention-guided decoder, effectively predicting lanes and minimizing false detections without requiring post-processing. LDNet leverages detailed event camera images to simplify full-resolution detections by extracting higher-dimensional features. The ASPP block expands the receptive field size without increasing training parameters, and the attention-guided decoder improves feature localization, eliminating the need for post-processing.
\\Liu et al. \cite{RN12} developed a deep learning-based autonomous lane-keeping system. They designed a robust lane detection and tracking system, introducing a lightweight network called LaneFCNet that optimized accuracy and processing time. This was followed by lane tracking to enhance detection performance and create an autonomous driving trajectory. Finally, they addressed the lane fitting problem using ridge regression, improving the overall system's effectiveness.
\section{Methodology}
The proposed lane detection algorithm leverages computer vision and deep learning techniques to accurately identify lane positions. It accounts for camera position variations and road surface imperfections, focusing on detecting and displaying deviations from the road center within a specified area. A dataset in MP4 video format from \cite{RN22} was utilized. During the initial pre-processing stage, the input video was decomposed into individual frames. Pre-processing involved using the Python library \textit{Albumentations}. This library converted each image from the BGR color space (used by OpenCV) to the RGB color space (preferred for deep learning models) and resized the images to 224x224 pixels. Various image augmentation methods were then applied, including rotation, additive noise, and adjustments to brightness and contrast. Specifically, the \textit{ShiftScaleRotate} transformation randomly rotated images, while the \textit{IAAAdditiveGaussianNoise} transformation added Gaussian noise, and the \textit{CLAHE} function improved image contrast.\textit{ RandomBrightness} and \textit{RandomGamma} functions were employed to alter image brightness and gamma, respectively, and the \textit{IASHarpen} function enhanced image sharpness. Blur effects were introduced using the Blur and \textit{MotionBlur} functions, while the \textit{RandomContrast} function adjusted image contrast, and the \textit{HueSaturationValue} function modified color, saturation, and value. Following these steps, a dataset comprising \textit{3075} images for training and \textit{129} images for validation was created. 
\\Two methods were used in the implementation. The FPN model was defined and implemented in the first method using the \textit{segmentation\_models\_pytorch} library. We used \textit{EfficientNet\-B0}, pre-trained by ImageNet, as an encoder in a FPN. In this way, \textit{EfficientNet\-B0} processes the input image and produces feature maps with different spatial resolutions (pyramidal levels). According to the hyperparameters mentioned in Table 1, \textit{Adam} optimizer function, \textit{softmax2d} activator function, and \textit{MultiDiceLoss} loss function are used to implement this model. This loss function was defined using a class for two right and left masks with \textit{BinaryDiceLossLeft}, and \textit{BinaryDiceLossRight} respectively. In addition, the learning rate was considered equal to \textit{1e-4}, and the model was implemented with \textit{4} epochs and \textit{8} batch sizes. Figure  \textit{1} shows an example of the image used to train the FPN model along with its corresponding label.

\begin{table}[width=.5\linewidth,cols=2,pos=h]
\caption{Hyperparameters used in the FPN model implementation.}\label{tbl1}
\begin{tabular*}{\tblwidth}{@{} CC@{} }
\toprule
Different hyperparameters of the model & Amount used \\
\midrule
Optimizer function & Adam  \\
Activator function & softmax2d  \\
Loss function & MultiDiceLoss  \\
Learning rate & 1e-4  \\
Number of epochs & 4  \\
Batch sizes & 8\\
\bottomrule
\end{tabular*}
\end{table}



\begin{figure}[h]
    \centering
    \includegraphics[width=0.5\textwidth]{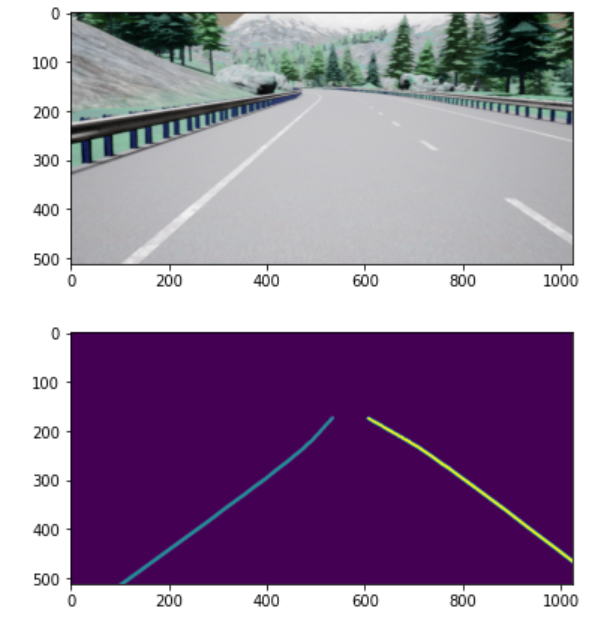} 
     \caption{An example of the image used to train the FPN model.}
     \label{fig:Figure1}
\end{figure}

In the second implementation, we used a U-Net model with attention blocks for image segmentation. This model takes an image of size \textit{256x320x3} as input. The encoder extracts features at multiple scales using convolutional layers and reduces spatial dimensions via max-pooling. In the decoder, spatial resolution is recovered through up-sampling, and attention blocks are applied to focus on relevant features. The attention layers are integrated into the upsampling path (decoder) during the skip connections. These attention layers refine the concatenation process between the upsampled decoder feature maps and the corresponding encoder feature maps, ensuring the model focuses on relevant features during the reconstruction of the high-resolution output \cite{li2022more}. This enhances the skip connections by concentrating on the most pertinent features from the encoder stages. The final output is a binary segmentation mask that highlights areas of interest in the input image. The attention mechanism significantly enhances the model's ability to focus on important parts of the image, improving segmentation accuracy. The model has \textit{15,117,829} parameters, indicating its complexity and capacity for detailed segmentation tasks. According to the hyperparameters mentioned in Table 2, the Adam optimizer function and binary cross entropy loss function were used in the implementation of this model. Also, the execution was done in 10 epochs with a batch size of 8. Figure \textit{2} shows an example of the image used for the attention-based U-Net model along with its label during training.

\begin{table}[width=.65\linewidth,cols=2,pos=h]
\caption{Hyperparameters used in the attention-based U-Net model implementation.}\label{tbl2}
\begin{tabular*}{\tblwidth}{@{} CC@{} }
\toprule
Different hyperparameters of the model & Amount used \\
\midrule
Optimizer function & Adam  \\
Loss function&Binary cross entropy \\
Number of epochs&10  \\
Batch sizes & 8\\
\bottomrule
\end{tabular*}
\end{table}

     
\begin{figure}[h]
    \centering
    \includegraphics[width=0.5\textwidth]{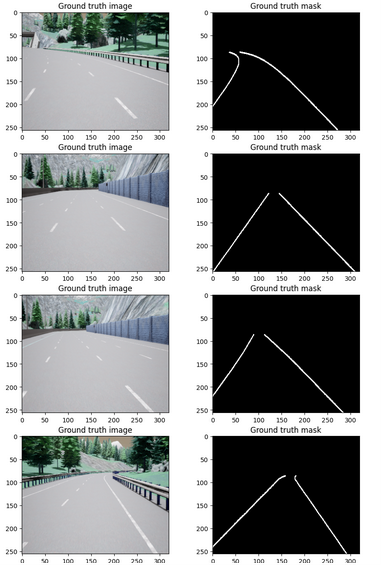} 
     \caption{Example of the images used for training the attention-based U-Net model.}
     \label{fig: Figure2}
\end{figure}
\section{Results and discussion}
Several evaluation criteria have been used in the evaluation of the two proposed methods. At first, the FPN model was evaluated using the Intersection over Union (IOU) criterion. Table 3 shows the evaluation results of the FPN model. According to this table, the total number of frames used in this implementation is 129 frames. After implementation, the model correctly recognized both lines in 113 frames with an accuracy of 87.59\%. Also, the model correctly recognizes only one line in 117 frames with an accuracy of 90.69\%.

\begin{table}[width=.7\linewidth,cols=2,pos=h]
\caption{Accuracy of the FPN model.}\label{tbl3}
\begin{tabular*}{\tblwidth}{@{} CC@{} }
\toprule
 & Value \\
\midrule
Total number of frames & 129 \\
The number of frames in which both lanes were correctly identified & 113 \\
Accuracy for both lanes correctly identified [\%] & 87.59\% \\
Number of frames in which at least one lane was correctly identified & 117\\
Accuracy for at least one lane was correctly identified [\%]	& 90.69\%\\
\bottomrule
\end{tabular*}
\end{table}



The output of the FPN model for the test data is shown in Figure 3. In this figure, the ground truth mask indicates the correct position of the right and left lanes. The two labels that predicted the left mask and predicted right mask also illustrate the positions predicted by the FPN model for the left and right lanes, respectively.
\begin{figure}[h]
    \centering
    \includegraphics[width=0.95\linewidth]{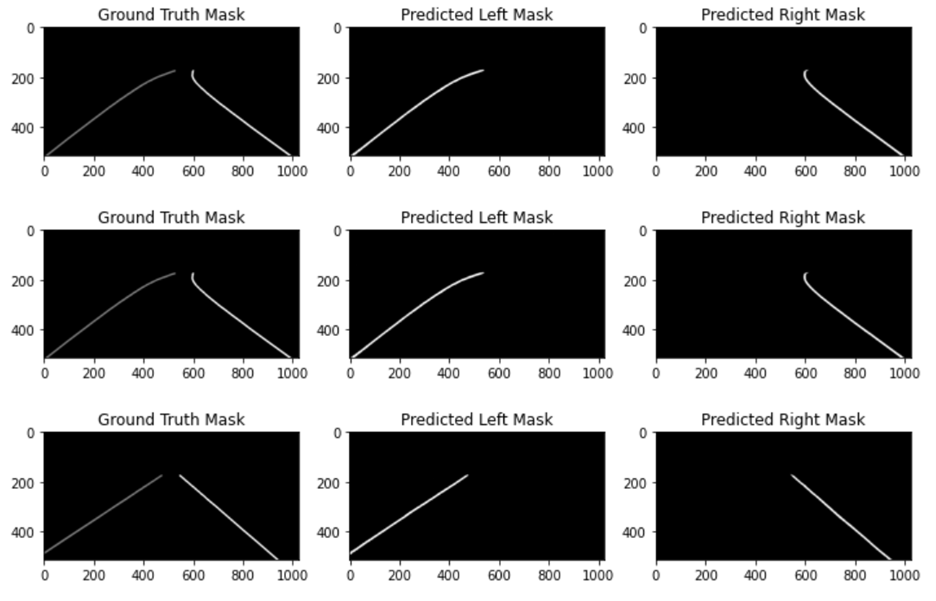}
    
    \caption{Examples of FPN model output for test data.}
    \label{fig: Figure3}
\end{figure}

Table 4 shows the evaluation results of this model for the test data. According to the results obtained from the implementation of this model, the accuracy, precision, recall, and IoU of this model were equal to 98.98\%, 81.50\%, 36.34\%, and 65.25\%, respectively.

\begin{table}[width=.5\linewidth,cols=2,pos=h]
\caption{Evaluation results of the attention-based U-Net model.}\label{tbl4}
\begin{tabular*}{\tblwidth}{@{} CC@{} }
\toprule
Evaluation criterion & Result \\
\midrule
Accuracy & 98.98\% \\
Precision & 81.50\%\\
Recall & 36.34\%\\
IoU & 65.25\%\\
\bottomrule
\end{tabular*}
\end{table}

To better compare the results of our proposed methods, we implemented the method mentioned in the paper by Brad et al. \cite{RN23}. Table 5 shows the results of this implementation. This method reached an accuracy of 98.59\% and the calculated IoU criterion was 49.29\%.

\begin{table}[width=.5\linewidth,cols=2,pos=h]
\caption{Implementation results of the Brad et al \cite{RN23} model.}\label{tbl5}
\begin{tabular*}{\tblwidth}{@{} CC@{} }
\toprule
Evaluation criterion & Result \\
\midrule
Accuracy & 98.59\% \\
IoU & 49.29\%\\
\bottomrule
\end{tabular*}
\end{table}
    

An analysis of the results presented in Tables 3, 4, and 5 reveals that our attention-based U-Net model outperforms the others in terms of accuracy. Additionally, it achieves a notable 15.96\% enhancement in the IoU criterion, setting a new standard in performance. Therefore, our models have superior performance compared to existing models.

\section{Conclusion}
Lane detection is crucial in autonomous vehicles and advanced driver assistance systems for identifying driving lane boundaries. The primary objective is to accurately determine the vehicle's position within the lane, which is essential for ensuring safe driving, lane keeping, and lane departure warning systems. The precision and reliability of lane detection systems are continually being enhanced through advanced algorithms and the integration of additional sensors like radar and lidar, providing a comprehensive understanding of the vehicle's surroundings. To this end, we proposed two innovative lane detection methods. The first method utilizes the FPN model, achieving a notable 87.59\% accuracy in lane detection. The second method enhances the U-Net model by incorporating attention layers in the decoder, resulting in an impressive accuracy of 98.98\% and an IoU of 65.25\%, which outperformed existing methods. This remarkable performance highlights our model's  precision in estimating road lane boundaries, setting a new benchmark in lane detection technology. 
\\Future research can focus on developing algorithms that perform robustly under challenging environmental conditions, such as low light, shadows, rain, snow, or blurred line markings. Additionally, it is feasible to advance beyond line detection to identify the types of lines, such as solid, dashed, and turn lanes, further enhancing the capabilities of lane detection systems.

\bibliographystyle{ieeetr}
\bibliography{References}

\vskip3pt

\bio{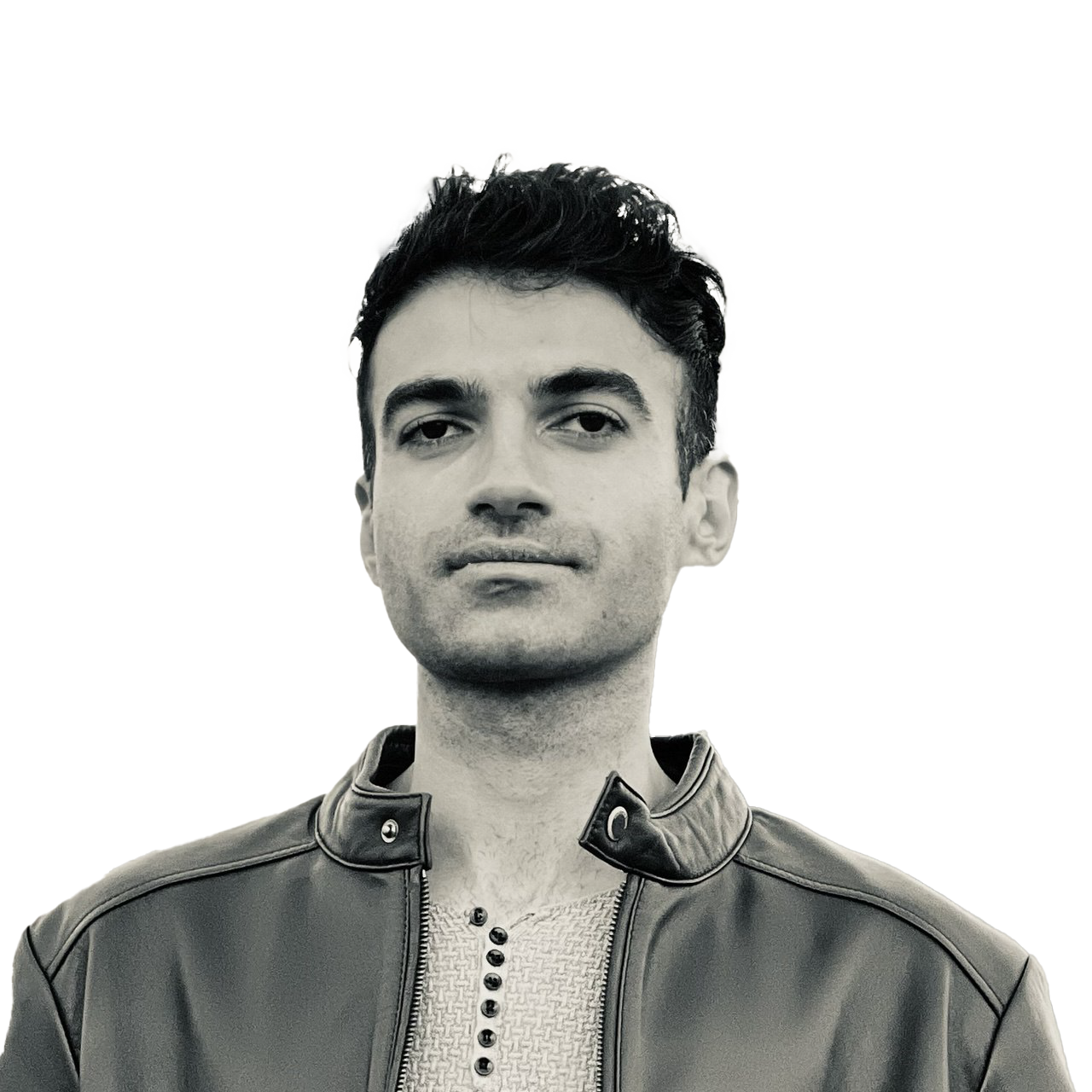}
Mohammadhamed Tangestanizade is a Ph.D. student at the University of California, Santa Cruz. He earned his Bachelor of Science degree in Electrical Engineering from Sharif University of Technology. Following this, he pursued and completed his Master’s degree at the University of California, Santa Cruz, in the Electrical and Computer Engineering (ECE) department. Currently, he is continuing his academic journey at the same institution, working towards a Ph.D. in Computer Science and Engineering (CSE). His research focuses on the development of deep learning and machine learning models, with various applications such as medical science data analysis and robotics.
\endbio

\bio{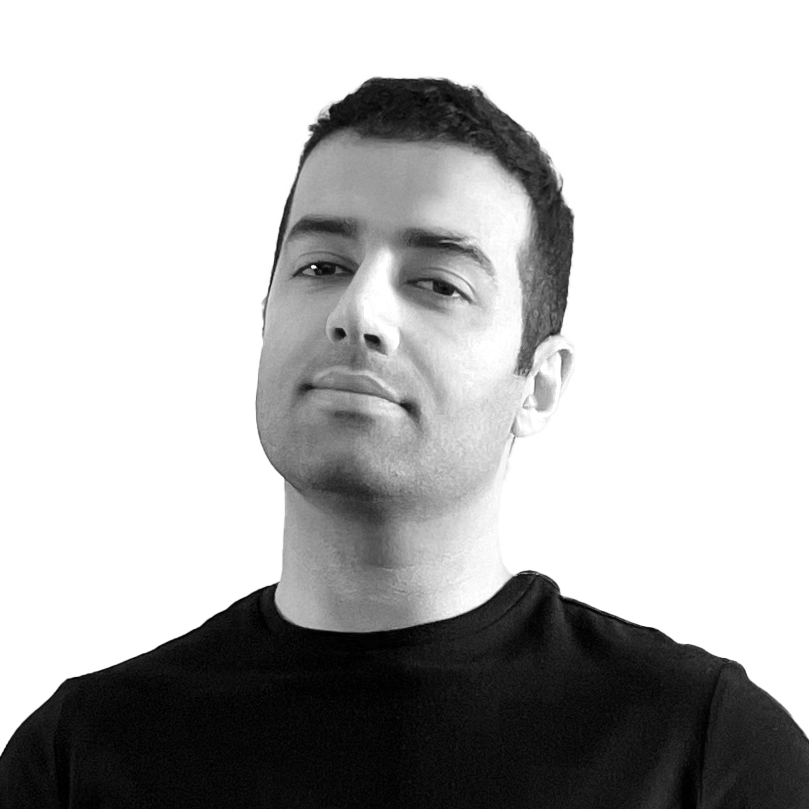}
Mohammad Dehghani Tezerjani is a PhD candidate at the University of North Texas, specializing in autonomous vehicles and robotics. His research focuses on vehicle control systems, motion planning, and perception, with a particular emphasis on integrating computer vision and machine learning techniques to enhance autonomous driving systems. Prior to his doctoral studies, he earned his Master's degree from Amirkabir University of Technology (Tehran Polytechnic), where he worked on advanced vehicle and mobile robots motion planning projects. His expertise spans multiple domains within autonomous systems, including sensor fusion, real-time perception, and the application of AI to complex decision-making in autonomous navigation.
\endbio

\bio{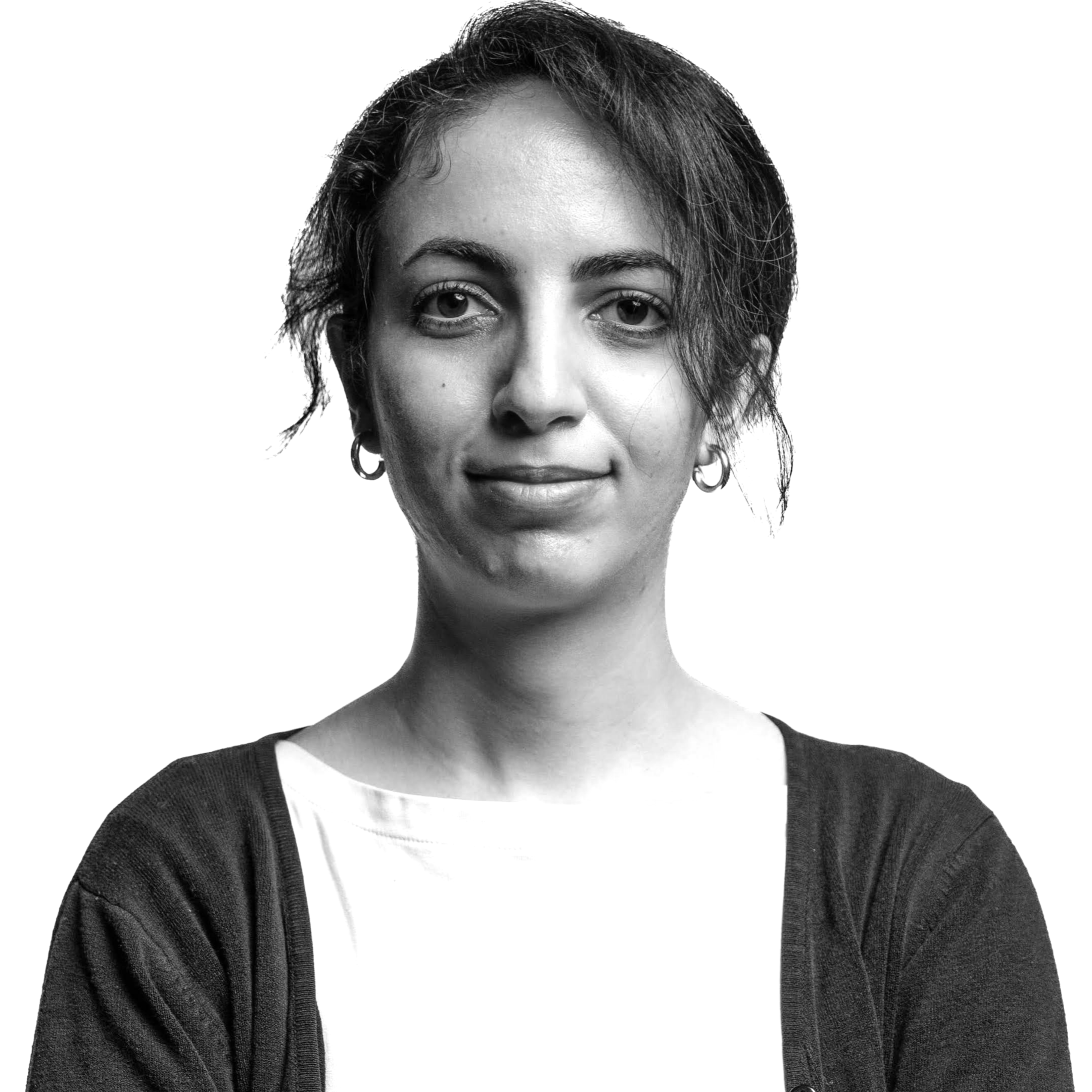}
Saba Yousefian Jazi is a PhD candidate in Computer Science and Engineering at the University of North Texas. She has been involved in various research projects, including non-marked tumor tracking in lung imaging, multi spectral image analysis, and point cloud clustering. Her most recent project focuses on developing machine learning and deep learning models and assessing their abilities in challenging tasks regarding data modalities such as images and video.
 \endbio

\end{document}